%% file: Document.tex
\documentclass[journal]{IEEEtran}

\usepackage{textcomp}

\usepackage{pifont}

\usepackage{tikz}
\usetikzlibrary{arrows, backgrounds, fit, shapes, shapes.callouts}
\usepackage{rotating}

\usepackage{tabu}

\usepackage{multirow}
\usepackage{longtable}
\usepackage{array}
\usepackage{adjustbox}

\usepackage{ctable}
\newcommand{\thline}{\specialrule{0.08em}{0em}{0em}} 

\usepackage[ruled,vlined,linesnumbered]{algorithm2e}

\usepackage[hidelinks]{hyperref}

\usepackage{graphicx}
\graphicspath{ {./Images/} }

\usepackage{subcaption}

\usepackage{multicol}

\usepackage{placeins}

\begin{document}

\title{Robust Speech-Workload Estimation for Intelligent Human-Robot Systems}
%
%
%

\author{Julian Fortune,
        ~Julie A. Adams,~\IEEEmembership{Senior Member,~IEEE},
        and~Jamison Heard,~\IEEEmembership{Member,~IEEE}
        }
        

\markboth{Submitted to an IEEE Journal}%
{Fortune \MakeLowercase{\textit{et al.}}:}

\maketitle

\begin{abstract}
\fontsize{9pt}{10pt}\selectfont
Demanding task environments (e.g., supervising a remotely piloted aircraft) require performing tasks quickly and accurately; however, periods of low and high operator workload can decrease task performance. Intelligent modulation of the system's demands and interaction modality in response to changes in operator workload state may increase performance by avoiding undesirable workload states. This system requires real-time estimation of each workload component (i.e., cognitive, physical, visual, speech, and auditory) to adapt the correct modality. Existing workload systems estimate multiple workload components post-hoc, but few estimate speech workload, or function in real-time. An algorithm to estimate speech workload and mitigate undesirable workload states in real-time is presented. An analysis of the algorithm's accuracy is presented, along with the results demonstrating the algorithm's generalizability across individuals and human-machine teaming paradigms. Real-time speech workload estimation is a crucial element towards developing adaptive human-machine systems.
\end{abstract}


%
\IEEEpeerreviewmaketitle



\input{1_Introduction}
\input{2_Related_Research}
\input{3_Algorithm_Design}

\input{4_Methodology}

\input{5_0_Environment}


\input{6_Conclusion}

\section{Acknowledgements}
\footnotesize
This work was supported by NASA Cooperative Agreement Number NNX16AB24A and by a Department of Defense Contract Number W81XWH-17-C-0252 from the Defense Medical Research and Development Program.

\bibliographystyle{ieeetr}
\fontsize{8pt}{9pt}\selectfont
\bibliography{Thesis,Heard}
\newpage

\newpage

\appendix
\normalsize

\input{appendix}

\end{document}

%% file: 1_Introduction.tex
\section{Introduction}

Humans teaming with robots in peer- or supervisory-based domains (e.g., nuclear power plant control room or civil support teams) \cite{Scholtz2003, Goodrich2007} need to achieve high task performance. Failure to do so may elicit a large financial cost, or the loss of life; thus, it is essential that the robot's interactions do not reduce team performance. Robust real-time multi-dimensional workload assessment algorithms may be leveraged to determine how a contemporaneous interaction occurs \cite{Heard2020CogSima}, as overall workload and its contributing components is related to task performance \cite{Wickens1984, mccracken1984analyses}. Performance decrements can occur when an individual is underloaded (too little workload) or overloaded (too much workload) subject to individual differences. However, typical algorithms classify cognitive workload solely and do not consider other workload components \cite{heard2018survey}, such as speech workload. Few algorithms exist that can classify speech workload in a manner that permits real-time deployment and generalizes across populations and task environments. 



Workload is defined as the ratio of resources demanded by a task to the resources a human has available to allocate to the task \cite{Wickens2003}. Overload occurs when a larger quantity of resources are required for a given task than the human is capable of dedicating to the task \cite{Wickens2003}, and can result in performance degradation. Conversely, an underload condition occurs when the task requires a small amount of resources, but the human has a large amount of available resources \cite{Wickens2003}. 
Individual human workload components can be assessed as part of comprehensive overall workload. McCracken and Aldrich \cite{mccracken1984analyses} decomposed workload into four workload components, the Visual, Auditory, Cognitive, and Psychomotor, in the VACP model. Mitchell \cite{mitchell2000} later expanded the VACP model by splitting psychomotor into two components: speech and motor. Speech workload is defined as any instance an individual is required to use their voice (simple and complex articulations) \cite{boles2007predicting}. Some definitions of speech workload include ``inner speech'', where an individual thinks about solutions (problem solving and critical thinking). This manuscript focuses solely on verbal speech, due to the complex nature of capturing/validating inner speech solely and that cognitive workload within the VACP model better captures these processes. Additionally, interpreting speech from another individual is captured within the auditory workload component, rather than the speech workload component.

Estimating each workload component is critical to providing a human-robot system with a robust understanding of the human's current available resources (workload). A previously developed multi-dimensional workload assessment algorithm \cite{Heard2019jcedm, Heard2019thri} has estimated each workload component; however, speech workload was estimated via an a priori workload model that does not capture individual differences. Thus, there is need to validate estimating speech workload via audio-based features. This work work developed and validated such an algorithm using spurious and non-spurious audio data from supervisory and peer human-robot teams. A prior version of this algorithm was used in real-time in an adaptive human-machine system \cite{Heard2020CogSima, fortune2020real, hfes}. The presented speech workload estimation approach expands the prior works significantly by improving the voice activity detection and voice intensity calculations, which reduces the overall computational complexity. Additionally, the algorithm's ability to generalize across human-robot teaming paradigms and across participant populations as analyzed.

%% file: 2_Related_Research.tex
\section{Related Research}

Speech-based feature extraction algorithms analyze speech audio data to estimate numerical values for certain speech characteristics (e.g., speech rate, pitch). Many of the objective speech workload metrics can be estimated using speech-based feature extraction algorithms that can form the basis for predicting speech workload, such as voice activity.

Voice activity is a binary classification between the presence or absence of human speech and enables the presented algorithm to avoid unnecessary calculations when there is no speech. Moattr and Homayounpour's \cite{simpleButEfficient} algorithm detected voice activity in real-time using short-term energy, spectral flatness, and the most dominant frequency in the power spectrum to discriminate between voiced and unvoiced audio segments. The algorithm adapted an energy threshold over time in order to distinguish regions potentially containing speech from silence, while accounting for recording differences and dynamic environments. The real-time algorithm was robust to noise, but was computationally expensive.

Speech workload estimation algorithms combine speech-based features intelligently. However, few approaches attempt to estimate speech workload; despite being a critical component to the overall workload state \cite{mitchell2000}. PHYSIOPRINT classified speech workload into \textit{Breath Hold}, \textit{Normal Breathing}, \textit{Short speech}, and \textit{Long speech} via an unspecified algorithm for processing speech audio and respiration rate data from physiological sensors \cite{popovic2015physioprint}. Similarly, Luig and Sontacchi \cite{luig2010workload} classified two speech workload levels using commonly used objective speech-workload metrics (e.g., formant locations, zero-crossing rate) for a single human-robot teaming domain.

More recent literature has focused on utilizing speech within a multi-modal algorithm to classify overall workload, such as combining speech and cardiovascular modalities \cite{magnusdottir2022assessing}. Deep-learning approaches (e.g., convolutional neural-networks) have also been employed to classify overall workload in real-time using speech-based measures \cite{sandoval2022real}. However, these multi-modal and deep-learning approaches do not provide diagnostic information about the human's overall workload state and suffer from individual differences \cite{magnusdottir2022assessing}.

Overall, few works focus solely on estimating speech-workload, as the works either classify a few speech-workload levels or use speech-based measures with other modalities to classify overall workload. Furthermore, the related literature does not focus on approaches that generalize between heterogeneous populations and human-robot teaming domains. The presented algorithm is the first approach that is capable providing a continuous speech-workload estimate that is robust to individual differences and can be employed in multiple human-robot teaming domains.

%% file: 3_Algorithm_Design.tex
\section{Speech Workload Estimation Algorithm} \label{algorithmDesign}

The speech workload estimation algorithm consists of two main components: feature extraction and learning. The feature extraction algorithms extract meaningful speech characteristics from the audio data and are used as inputs for making predictions with the learning algorithm. The presented algorithm improves significantly on a prior algorithm version \cite{hfes}. The presented method incorporates an auto-correlation method (Section \ref{algorithmDesignPitch}) to estimate pitch, instead of the loudest component of the power spectrum \cite{hfes}. Performance enhancements to the voice activity and syllable extraction algorithms (Sections \ref{algorithmDesignVoiceActivity} and \ref{algorithmDesignSyllables}, respectively) are incorporated. Validation of each metric is presented in the Appendix Section \ref{appendix:feats}. 




\vspace{-0.5cm}
\subsection{Metrics} \label{algorithmDesignMetrics}

A captured audio stream is converted from stereo (2-channel) to mono-channel audio data using the Wavio library \cite{wavio}. Feature extraction is performed using 5-second \emph{windows} of audio with a step size of 1 second. Complex speech characteristics, consisting of voice intensity, pitch, voice activity, and syllables; are extracted from each window using the corresponding feature extraction algorithms. These algorithms perform calculations on audio \emph{frames} that are at 10 millisecond (ms) intervals looking back over 50 ms of audio, where empirical testing determined these to values to work well. Note that audio frames are distinct from the \emph{windows} of input audio data. This process produces arrays of values, where each element corresponds to a frame. The resulting arrays are utilized when one feature extraction algorithm depends on the data from another algorithm. Descriptive statistics (e.g., mean, standard deviation (Std. Dev.)) are calculated from each array and fed into a neural network for the final estimate.

\subsubsection{Intensity}  \label{algorithmDesignEnergy}

Audio \emph{intensity} is calculated using root-mean-square for each frame. 
The mean and standard deviation are calculated across all frames to generate representative intensity values for the window. The intensity algorithm iterates through each sample in a linear fashion in linear (O(n)) run time.


\subsubsection{Pitch} \label{algorithmDesignPitch}

\emph{Pitch} is estimated for each frame using an auto-correlation method implemented by the speech-analysis application PRAAT \cite{praat} accessed via the Parselmouth library \cite{parselmouth}. Auto-correlation shifts the signal by small distances and calculates the correlation between the original signal and the shifted signal. The shift that produces the largest correlation corresponds to the estimated wavelength, and the fundamental frequency (pitch) is calculated based on the selected wavelength. Frequencies above 400 Hz are rejected, as this threshold is well above the upper range of human speech (320 Hz). Frames are set to 0 if above a silence threshold calculated by multiplying the ambient noise level by a signal to noise ratio of 4 determined via testing on speech recordings. The ambient noise level is the $10^{th}$ percentile of the entire window's mean intensity. Audio frames with fewer than four consecutive non-zero values are assumed to be non-speech noises, and are removed.

The pitch values are used by the other feature extraction algorithms (i.e., voice activity), but the final pitch values are produced after all other features were calculated. Pitch values not within 100 ms of a voice activity value of \texttt{true} (e.g., a positive speech identification) are discarded; thus, rejecting extraneous sounds. The pitch values' mean and standard deviation are calculated in linear (O(n)) time to produce representative values for the audio processing window. 

\subsubsection{Voice activity}  \label{algorithmDesignVoiceActivity}

The \emph{voice activity} algorithm uses short-term energy, zero-crossing rate, and pitch to classify voice activity by identifying sufficiently loud regions with zero-crossing rates and pitches in the human vocal range \cite{simpleButEfficient}. Short term energy is calculated as $\mathrm{Energy} = \sum_{t=0}^{N-1} s^2(t)$ \cite{nandhini2014shortterm}, where $s(t)$ is the discrete value of the signal in sample $t$, and $N$ is the total number of samples in the audio data. The zero-crossing rate is produced by the Librosa library \cite{librosa}.

\input{Algorithms/voiceActivity}

The energy threshold is set via an adaptive process \cite{simpleButEfficient} (see Algorithm \ref{adaptiveEnergyAlgorithm}). The \texttt{primaryThreshold} parameter is 40 dB \cite{simpleButEfficient} (line 1). The variable \texttt{minEnergy} is determined by averaging the energy from the first 30 audio samples, which are assumed to be silent (line 2). The energy threshold is initialized to $\texttt{primaryThreshold} \times \log_e$ \texttt{minEnergy} (line 3). The energy, zero-crossing rate, and pitch are calculated for the audio, and stored in variables (lines 6--8). The \texttt{voiceActivity} array stores the identification of voice as a Boolean value is initialized to \texttt{false} (line 10).

The algorithm iterates through every frame (lines 11--19) and dynamically sets the \texttt{energyThreshold} (line 12). The maximum and minimum thresholds for zero-crossing rate are 0.04 and 0.008 crossings per sample, respectively; and were chosen based on human vocal ranges (line 12). The pitch threshold is any non-zero value (line 12), because the simple voice detection criterion rejects non-voiced pitches. The algorithm examines all pitch values within eight frames, on either side of the current frame, as the pitch algorithm is less tolerant to noise than is desirable in voice activity detection (line 12). If the thresholds for speech are met, the element at the current frame index in \texttt{voiceActivity} is set to \texttt{true} to indicate the presence of voice (lines 12--13). If the thresholds for speech are not met, \texttt{silenceCount} is incremented by 1 (line 15). The variable \texttt{minEnergy} is updated to $\frac{ ( \texttt{silenceCount} \times \texttt{minEnergy} ) + \texttt{energy}}{\texttt{silenceCount} - 1} $, where \texttt{energy} is the frame energy (line 17). The energy threshold is updated to $\texttt{primaryThreshold} \times \log_e \texttt{minEnergy}$ \cite{simpleButEfficient}.
 
Several post-processing steps occur after the loop terminates. Consecutive voice activity values shorter than 10 frames (100 ms) are assumed to be erroneous and were removed (line 20). Ten frames was chosen as the minimum length in order to minimize false alarms when testing on validation files. The \texttt{true} and \texttt{false} values are converted to 1 and 0, respectively (line 21) prior to calculating the mean and standard deviation (line 22). The voice activity algorithm is based on features with linear (O(n)) run time, and iterated through the audio data linearly, resulting in a linear run time.

\subsubsection{Syllables} \label{algorithmDesignSyllables}

The \emph{syllable} extraction process uses intensity (Section \ref{algorithmDesignEnergy}), pitch (Section \ref{algorithmDesignPitch}), and zero-crossing rate. Voiced peaks in intensity are considered to be syllables. The algorithm identifies intensity peaks for the audio data using the SciPy Signal library \cite{SciPy2020}. The process identifies peaks at least two frames wide, and at least four frames apart \cite{SciPy2020}. A peak is marked as a syllable if two conditions are met. If the frame's zero-crossing rate value is less than 0.06 chosen based on human vocal ranges, and there is a non-zero pitch value within four frames, before or after. Pitch is used to verify that a peak was voiced. If the peak is a plosive (a consonant-based burst of noise), the margin of four frames verifies that voiced audio follows shortly after. The number of syllables is divided by the length of the audio data in order to derive the speaking rate in syllables per second, which requires linear (O(n)) time.

\subsection{Machine Learning} \label{learning}

A speech workload estimate is produced using a five-layer neural network, which is the same architecture used in previous works \cite{Heard2019thri, Heard2019jcedm}. The input layer consists of seven neurons, corresponding to the seven extracted metrics: speaking rate as well as mean and Std. Dev. for speech intensity, pitch, and voice activity. The three hidden layers each have 256 neurons each with a rectified linear-unit activation function. The output layer consists of a single neuron, corresponding to the speech workload estimate. The ADAM optimizer \cite{kingma2014adam} is used for training, with an initial learning rate of 0.001 and batch size of 64. The ground-truth labels used for training and validation ranged from 0--4 and were produced using IMPRINT Pro, a human factors workload modeling tool \cite{archer2005imprint}.

\subsection{Speech Workload Estimation Algorithm}

The speech workload estimation algorithm (\ref{completeAlgorithm}), can estimate speech-workload in real-time; thus, \emph{windows} of audio are processed (lines 1--2), either from an audio stream captured in real-time, or from post-hoc file analyses. The algorithm transforms the audio signal from a stereo signal to a mono signal, if needed (line 3) and calculates the voice activity (Section \ref{algorithmDesignVoiceActivity}) for each window (line 4) in order to determine if speech is present. If the voice activity value is 0, the algorithm returns that value (line 18). Otherwise, the algorithm extracts features (lines 7--11), inputs those features to the neural network (lines 13--15), and returns a non-zero estimate.

\input{Algorithms/speechWorkloadEstimation} 

%% file: Algorithms/voiceActivity.tex
\begin{algorithm}[h]
\SetAlgoLined
\footnotesize
\DontPrintSemicolon
\KwResult{Tuple of floating-point values [0,1] (mean, st. dev)}

primaryThreshold $\leftarrow$ 40\;
minEnergy $\leftarrow$ mean(calculateEnergy(data: audio[0:30]))\;
energyThreshold $\leftarrow$ primaryThreshold * log(minEnergy)\;
silenceCount $\leftarrow$ 0\;
\;
energyArray $\leftarrow$ calculateEnergy(data: audio, step: 10, size: 50)\;
zcrArray $\leftarrow$ calculateZeroCrossingRate(data: audio, step: 10, size: 50)\;
pitchArray $\leftarrow$ calculatePitch(data: audio, step: 10, size: 50)\;
\;
voiceActivity $\leftarrow$ [False] * length(energy)\;
\For{i in range(0, length(energy))}{
    \eIf {energyArray[i] $>$  energyThreshold \textbf{and} 
          0.04 $<$ zcrArray[i] $<$ 0.008 \textbf{and}
          nonZeroValue(data: pitchArray, within: 8, ofIndex: i)} {
        voiceActivity[i] $\leftarrow$ True\;
    } {
       silenceCount += 1\;
    }
    
    minEnergy $\leftarrow$ ((silenceCount * minEnergy) + energyArray[i]) / (silenceCount+1)\;
    energyThreshold $\leftarrow$ primaryThreshold * log(minEnergy)\;
}
removeRuns(data: voiceActivity, lessThan: 10)\;
voiceActivity $\leftarrow$ convertBooleanToInteger(voiceActivity)\;
return mean(voiceActivity), stDev(voiceActivity)\;

\caption{Voice Activity Detection Algorithm.}
\label{adaptiveEnergyAlgorithm}

\end{algorithm}

%% file: Algorithms/speechWorkloadEstimation.tex
\begin{algorithm}[h]
\SetAlgoLined
\footnotesize
\DontPrintSemicolon
\KwResult{Floating-point value [0,4]}
\For{i $\leftarrow$ 0, i $<$ length(audio), i $\leftarrow$ i + stepSize}{
    currentAudio $\leftarrow$ audio[i:i$+$windowSize]\;
    currentAudio.convertToMono()\;
    voiceActivityMean, voiceActivityStDev $\leftarrow$ getVoiceActivity(currentAudio)\;
    \;
    \eIf{voiceActivityMean $>$ 0}{
        intensityMean, intensityStDev $\leftarrow$ getIntensity(currentAudio)\;
        pitchMean, pitchStDev $\leftarrow$ getPitch(currentAudio)\;
        syllablesPerSecond $\leftarrow$ getSyllables(currentAudio)\;
        \;
        features $\leftarrow$ [intensityMean, intensityStDev, pitchMean, pitchStDev, voiceActivityMean, voiceActivityStDev, syllablesPerSecond]\;
        \;
        estimate $\leftarrow$ neuralNetwork(features)\;
        return estimate\;
    } {
    return 0\;
    }
 }
 
\caption{Speech Workload Estimation}
\label{completeAlgorithm}
\end{algorithm}

%% file: 4_Methodology.tex
\section{Human Subjects Evaluation Methodology}

The speech data, physiological metrics, and workload values used in algorithm's experimental validation were collected during two human subjects evaluations \cite{heard2019dissertation}: Supervisory and Peer. These evaluations focused on the overall workload estimation algorithm \cite{heard2019diagnostic}, and were not conducted solely for the purpose of developing the presented algorithm. Each evaluation's task was modeled using IMPRINT Pro \cite{archer2005imprint, heard2019dissertation} to ensure that the independent variable (workload) varied between experimental conditions. Both evaluations were approved by an appropriate Institutional Review Board.

\subsection{Supervisory Evaluation} \label{supervisoryEval}

The Supervisory Evaluation validated workload assessment algorithms in a multi-tasking supervisory-based human-machine teaming scenario \cite{Heard2019thri}. The evaluation spanned two days: the first day comprised of one trial for each workload state (underload (UL), normal load (NL), and overload (OL)) and the second day comprised a single trial, where participants' workload levels transitioned at predetermined times in order to emulate real-world conditions. The participants completed a consent form and demographic questionnaire upon arrival on the first day. The participants were outfitted on both days with: a BIOPAC\textsuperscript{\textregistered} Bioharness\textsuperscript{TM} 3 \cite{biopac} to capture the participants' physiological metrics 
and a Shure microphone. 

The participants completed a 10-minute training session during the first day followed by three 15-minute trials, with a 5-minute break after the training session and between each trial (necessary to allow workload levels to return to their resting level \cite{reimer2009road}). The three counterbalanced trials corresponded to the three workload states: UL, NL, OL. The second day involved a single 35-minute trial comprised of seven 5-minute workload conditions, with one of three orderings: 
\begin{itemize}
	\item UL-NL-OL-UL-OL-NL-UL
	\item NL-OL-UL-OL-NL-UL-NL
	\item OL-UL-OL-NL-UL-NL-OL.
\end{itemize}
The orderings where chosen, such that each workload transition occurred once. The 5-minute duration per condition was chosen to reflect the time needed for physiological signals to respond to the workload transition \cite{Castor2003}.



\begin{figure}[h]
\centering
\includegraphics[width=250pt]{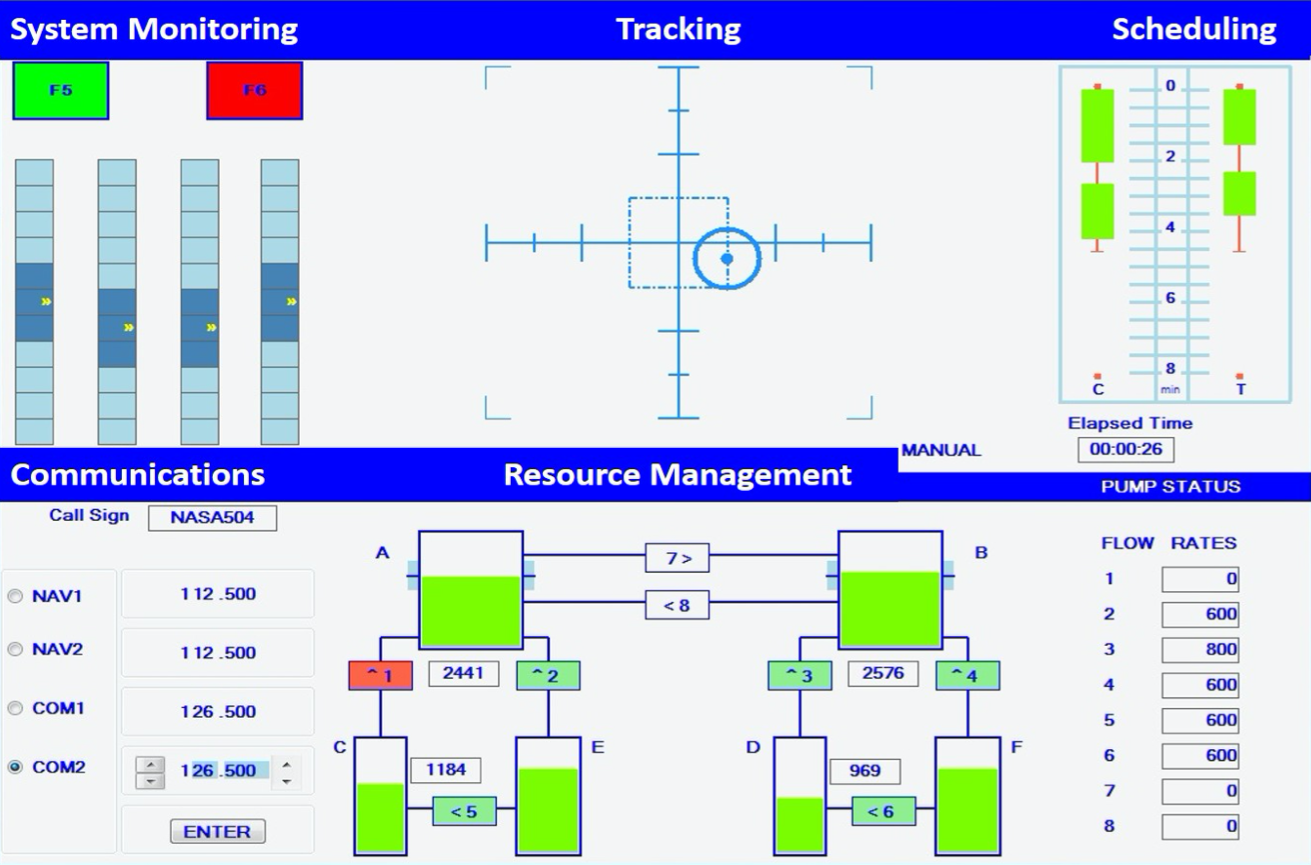}
\caption{ The NASA MATB-II Task Environment.}
\label{nasaMATBScreenshot}

\end{figure}

The evaluation utilized the NASA MATB-II (depicted in Fig. \ref{nasaMATBScreenshot}), which incorporates four concurrent unmanned aerial vehicle tasks: tracking, system monitoring, resource management, and communication monitoring. 
The tracking task (top center of Figure \ref{nasaMATBScreenshot}) required participants to center a randomly-drifting representation of an unmanned aircraft's path (the dot) in a two-dimensional cross-hairs. 
The task had two modes: automatic and manual. The participant physically controlled the aircraft's flight using the joystick in manual mode. The software automatically corrected the aircraft's position so that it remained in the center, without participant intervention, when in the automatic mode. The task was set to automatic mode during the UL condition, and manual mode during the OL condition The tracking mode alternated between automatic and manual mode every 2.5 minutes in the NL condition.

The system monitoring task (top left of Figure \ref{nasaMATBScreenshot}) required monitoring two colored buttons and four gauges. The first button was green when in range, and grey when out of range, while the second button was grey when in range, and red when out of range. The participant was to click on an out of range button in order to reset it. The four gauges fluctuated up and down randomly, typically hovering near the middle.  A gauge was out of range when the indicator (yellow triangle) was either too high or too low, and the acceptable range of readings was marked by the dark region of the gauge. The participants were to click on an out-of-range gauge in order to reset the gauge. The UL condition had a single out-of-range instance per minute, the NL condition had five instances per minute, and the OL condition had twenty instances per minute.

The resource management task (bottom center of Fig. \ref{nasaMATBScreenshot}) included six fuel tanks (A-F) and eight fuel pumps. Each fuel pump moved fuel from one tank to another tank, and the direction was indicated by an arrow. Participants were to maintain the fuel levels in Tanks A and B by turning individual pumps on and off. Tanks C and D had a finite fuel level, but Tanks E and F had an infinite fuel supply. A pump failure, indicated by the pump turning red, was possible, and resulted in the pump ceasing to move fuel. Zero pump failures occurred in the UL condition, while two or more pumps failed during the OL condition. The NL condition alternated between zero pumps failing and one to two pumps failing every minute.

The communications task (bottom left of Figure \ref{nasaMATBScreenshot}) involved listening continuously to air-traffic control request for radio changes. Request were of the form: ``NASA 504, please change your COM 1 radio to frequency 127.550.'' The participants were to change the specified radio to the specified frequency via clicking arrows when the request was directed to ``NASA 504'', while all other requests were to be ignored. The original communications task required no speech, but a required verbal response was added. Responses were to be of the form: ``This is NASA 504 tuning my COM 1 radio to frequency 127.550.'' The UL condition contained fewer than two requests during the condition, the NL condition contained two to eight requests, and the OL condition contained eight to twelve requests.


The objective dependent metrics included physiological data (e.g., heart rate, heart rate variability, skin temperature, respiration rate, and posture data), speech audio data, and performance measures automatically collected by the NASA MATB-II system. The subjective metrics included in-situ workload ratings \cite{harriot2013peer} and the NASA Task Load Index \cite{hart1988NASATLX}. The in-situ ratings were administered three times during each workload condition on the first evaluation day, during which participants were asked to verbally rate their cognitive, motor, tactile, auditory, speech, and visual workload levels using a Likert scale of 1--5. The same ratings were administered on the second evaluation day, but at five-minute intervals. The NASA-TLX was administered at the end of each trial. 


Thirty participants completed the evaluation (18 females and 12 males). Participants' ages ranged from 18 to 62, with an average of 25.70 years (Std. Dev. = 8.65).

\subsection{Peer-based Evaluation} \label{peerBasedEval} 

The Peer-based Evaluation analyzed differences in workload and task performance in human-robot peer-based teams \cite{harriot2013peer, heard2019diagnostic, Heard2019thri}. The participants completed four sequential Civil Support team disaster response tasks with a robot, where each task was modeled using IMPRINT Pro. Participants were assigned to either a \textit{low} or \textit{high} workload condition for each of the four tasks. A Pioneer 3-DX robot served as the robot assistant. An experimenter controlled the robot's movements and the robot's speech was generated based on prespecified or adapted in real-time scripts, but participants were told the robot's movements and speech were autonomous.

The task environment modeled a disaster scenario\footnote{Please see Appendix Section \ref{appendix:peer} for images depicting the tasks.}, and the tasks were ordered approximately as they are completed by civil support teams. The fifteen-minute tasks were: (1) identify suspicious items from images, (2) search for hazardous materials, and collect samples of both (3) solid and (4) liquid hazardous materials. All participants completed the tasks in the same order. Workload was manipulated to be NL or OL. Each participant was assigned a random workload condition for each task. All tasks were completed in offices, hallways, and laboratories in an academic building.

The participants completed a demographic questionnaire, received a task briefing, and were shown a training video before being outfitted with a BIOPAC\textsuperscript{\textregistered} Bioharness\textsuperscript{TM} 3, a Scosche Rhythm+ monitor, a Fitbit activity monitor, a camera, a walkie-talkie, a Shure microphone headset, a reflective yellow vest, goggles, a face mask, and gloves. A ten pound backpack was worn during the second and fourth tasks, simulating the personal protective gear worn by support teams.

The first task involved identifying suspicious items in digital images and annotating the items with text or circles. The participants, using a Google Nexus 7 tablet, received folders containing three images to search at predetermined times during the task. The robot produced an audible beep when a folder was received. The robot verbally asked the participant guiding questions and offered relevant information gained from air samples and notes related to the image under investigation. Two folders of images were received during the NL condition, and four folders during the OL workload condition.

The second task required physically performing an exhaustive search of a hallway environment in order to identify hazardous materials. The participant's role was to locate and photograph suspicious items that were located above the robot's field of view or that the robot was unable to manipulate. The robot scanned the floor for items, collected air samples, and, at predetermined times (3:45, 7:30, 11:15, and 15:00), produced an audible beep when the participant needed to send the current information to incident command. The robot spoke to the participant in order to discuss the nature of items found, and to advise the participant to investigate an item requiring attention, if the participant had overlooked it. Four hazardous materials were present in the NL condition, and eight items were present in the OL condition.

Tasks three and four involved collecting samples of solid (colored sand) and liquid (dyed water) contaminants, respectively. The participants collected the samples while wearing gloves under the guidance of auditory instructions from the robot. Two samples were collected in the NL condition, and four samples were collected in the OL condition.



The dependent variables consisted of objective metrics and subjective ratings. The objective metrics included physiological responses (the same as Section \ref{supervisoryEval}), speech audio data, and performance measures (e.g., primary task response time). The subjective metrics included the in-situ workload ratings and a post-trial NASA-TLX workload rating.


Eighteen participants completed the evaluation (9 male and 9 female). Participant's ages ranged from 18 to 38, with an average age of 22 years (Std. Dev. = 5.5).

%% file: 5_0_Environment.tex
\section{Results}

\tabulinesep=1mm

\newcommand{\h}[2]{$ \textbf{H}^{\textrm{\textbf{\scriptsize#2}}}
                               _{\textrm{\textbf{\scriptsize#1}}} $}

\newcommand{\R}[1]{\textbf{R\textsubscript{#1}}}


\newcommand{\allCond}{All} 
\newcommand{\bothCond}{All} 

\newcommand{\realWorldQuestion}{Can the algorithm estimate accurately an individual's speech workload level solely using audio data?}
\newcommand{\popGenQuestion}{Can the algorithm estimate accurately an arbitrary, unseen individual's speech workload level?}
\newcommand{\parGenQuestion}{Can the algorithm estimate accurately an individual's speech workload level in multiple human-robot teaming paradigms?}
\newcommand{\envGenQuestion}{Can the algorithm estimate accurately an arbitrary, unseen individual's speech workload level in multiple task environments solely using audio data?}
\newcommand{\realTimeQuestion}{What is the optimal window size for real-time speech workload estimation?}
\newcommand{\featureSetQuestion}[1]{Does adding respiration rate and filler utterance features to the base set of features#1 improve speech workload estimation accuracy?}

\newcommand{\descriptiveCaption}{Descriptive Statistics for the Algorithm Estimates and IMPRINT Pro Model Speech Workload Predictions by Data Set and Workload Condition.}
\newcommand{\descriptiveCaptionNote}{\descriptiveCaption \textbf{Note:} All the minimums are 0 and are not shown.}
\newcommand{\correlationRMSECaption}{Correlation, RMSE, and Percent Error Between the Algorithm Estimates and IMPRINT Pro Model Speech Workload Predictions by Data Set and Workload Condition.}
\newcommand{\correlationRMSECaptionNote}{\correlationRMSECaption \textbf{Note:} ** represents p $<$ 0.0001 and * represents p $<$ 0.05.}
\newcommand{\rmseCaption}{}

\input{5_0_Introduction}

\input{5_1}

\input{5_2}

\input{5_3}
\input{5_6}
\input{5_Discussion}

%% file: 5_0_Introduction.tex


Three research questions guided the analysis. The first research question (\R{1}) was: \emph{\realWorldQuestion}  The algorithm needs to estimate workload accurately when trained and tested on audio data from the same individuals. 
The algorithm also needs to estimate workload for individuals on which it has not been trained, in order to support applications where it is not practical to collect training data from all users. This leads to the research question (\R{2}): \emph{\popGenQuestion} Real-world applications will involve deployments in circumstances that differ completely from those in which training occurred. 
The algorithm's generalizability across human-robot teaming paradigms is assessed by the research question (\R{3}): \emph{\parGenQuestion}

Three post-hoc experiments, corresponding to each research question, analyzed the performance of the algorithm. All experiments used a window size of five seconds for extracting the features. The training epoch size was 50 for the Population Generalizability experiment, in order to decrease the analysis run-time, and 100 for all other experiments. 

IMPRINT Pro modeled overall workload and its contributing components a priori for each evaluation using task graphs that are anchored to specific demands (e.g., simple speech is anchored to a value of 2.0) \cite{heard2019dissertation}. These models serve as a continuous diagnostic representation of a human's overall workload state and were used as the ``ground-truth'' labels for algorithm training. However, these models are subject to synchronization errors and participant variability. For example, the supervisory evaluation incorporates air-traffic control requests that a participant audibly responds to. IMPRINT Pro modeled a participant's speech workload when a response was expected; however, a participant may have a delayed response or even fail to respond. Furthermore, participants may speak sparsely in an unpredictable manner, such as thinking out loud. This process creates noisy labels for algorithm training and validation, which makes it difficult to discern whether errors resulted from the algorithm or inaccurate labels.  Two datasets were created for each analysis to provide a more robust representation of the results: \emph{unfiltered} and \emph{filtered}. The \emph{unfiltered} dataset used the algorithm's estimates and IMPRINT Pro models' predictions from the entire training/testing dataset. The \emph{filtered} dataset selected instances where the speech workload models agreed with the algorithm's voice activity detection (Section \ref{algorithmDesignVoiceActivity}) for algorithm validation. \emph{Agreement} occurred when the IMPRINT Pro value was zero and the voice activity value was \texttt{False}, or the IMPRINT Pro value was non-zero and the voice activity value was \texttt{True}. Although this process artificially inflates the correlation values, it provides a more accurate depiction of algorithm estimation error. Thus, correlation values are not reported for the \emph{filtered} dataset. The descriptive statistics for each analysis are provided in the Appendix Section \ref{appendix:stats}, while Appendix Section \ref{appendix:run} depicts the algorithm run-time for various window sizes.




%% file: 5_1.tex

\subsection{Emulated Real-World Conditions} \label{emulatedRealWorld}

This experiment analyzed the accuracy of the speech workload estimation algorithm on individuals (\R{1}) when trained on separate workload conditions and tested in emulated-conditions, where speech workload fluctuates. The algorithm was trained on all participant data collected during the first Supervisory Evaluation day (Section \ref{supervisoryEval}), and tested on all participant data captured on the second day. Overall descriptive statistics can be found in the appendix.

The first hypothesis focuses on estimation error stating that the root-mean-square-error (RMSE) between the algorithm's estimates and the model predictions will be less than 5\%. RMSE does not provide temporal alignment information; thus \h{PC}{1} predicted that the Pearson's correlation coefficient between the IMPRINT Pro model's predictions and speech workload estimates will be positive and significant.

\subsubsection{Results}


The RMSE between the algorithm's estimates and the IMPRINT Pro predictions are provided in Table \ref{realWorldCorrelationRMSE}.
The RMSE value as a percent of the model's predicted mean, labeled ``Percent Error'', is included for each dataset and condition as a measure of the mean error's scale in the context of the speech workload estimates. The percent errors in all conditions and overall for the unfiltered dataset are above 200\%, a magnitude of two times the mean IMPRINT Pro workload prediction, which indicates that the algorithm's estimates severely disagreed with the IMPRINT Pro predictions. The percent error is 0.219\% and 0.424\% in the OL and NL conditions, respectively, for the filtered dataset. 


The Pearson's Correlation Coefficient between the algorithm's estimates and the speech workload model predictions from the filtered and unfiltered datasets are presented in Table \ref{realWorldCorrelationRMSE} by workload condition.
The correlation coefficients for the unfiltered data are all positive. The OL condition in the unfiltered dataset was not significant, but all remaining correlation coefficients were highly significant.



\begin{table}[!htb]
\caption{Correlation, RMSE, and Percent Error between the
algorithm estimates and the model predictions by Data Set and Workload Condition. \textbf{Note:} ** represents p $<$ 0.0001.} 
\label{realWorldCorrelationRMSE}
\input{Results/realWorld/correlationRMSE}

\end{table}

\subsubsection{Discussion}

Demonstrating the speech workload estimation algorithm's ability to perform well when trained and tested on data from the same individuals using only audio data is critical to understanding algorithm error without individual differences confounds. 
The RMSE determined the magnitude of the error between the algorithm estimates and IMPRINT Pro, but the results did not support \h{RMSE}{1} fully. The high error in the unfiltered dataset is due to the IMPRINT Pro model not accounting for variability in the timing of participants' vocal responses to prompts. The error in the filtered, UL condition is likely due to an algorithm bias that may result from the algorithm treating simple speech as complex speech. 

The algorithm performed well on a filtered dataset when trained and tested on the same individuals. Some domains may not provide prior data for the individuals to be assessed, or it may be infeasible to collect data from all individuals before estimating workload. The next experiment provides insight into the algorithm's performance when it is tested on individuals excluded from the training data set.


The correlation determined how well the algorithm tracked speech workload changes. \h{PC}{1} stated that the correlation coefficients between the IMPRINT Pro model predictions and the speech workload algorithm estimates will be positive and significant, which was almost entirely supported. The positive coefficients demonstrate that the algorithm estimates generally respond to changes in workload in the same direction as the IMPRINT Pro predictions. The coefficient magnitudes in the filtered dataset show that the algorithm estimates are highly aligned to the IMPRINT Pro predictions for all workload conditions, and that the estimates track extremely well overall.

The lower correlations with the unfiltered dataset show that the algorithm estimates tracked the model predictions poorly, which may be a result of deficiencies in the IMPRINT Pro model. IMPRINT Pro assumed that participants began speaking at a specific time for a specific duration in response to speech prompts (e.g., radio communications). However, participants' response times were inconsistent, participants spoke for varying durations, and some participants missed communication events entirely. The IMPRINT Pro models do not adequately represent the human's actual speech workload.

%% file: Results/realWorld/correlationRMSE.tex

\centering
\begin{tabu}{cc|llr}
\thline
\textbf{Dataset}               & \textbf{Condition} &   \textbf{Correlation} &   \textbf{RMSE} &   \textbf{Percent Error} \\
\hline
\multirow[c]{4}{*}{Unfiltered} &                 UL & 0.144** & 1.238 & 242.238\% \\
                               &                 NL & 0.046** & 1.819 & 358.997\% \\
                               &                 OL & 0.008 & 2.494 & 248.383\% \\
                               \cline{2-4}
                               &           \allCond & 0.088** & 1.859 & 279.272\% \\
\hline
\multirow[c]{4}{*}{Filtered}   &                 UL & - & 1.295 & 154.95\% \\
                               &                 NL & -& 0.005 & 0.388\% \\
                               &                 OL & -& 0.006 & 0.212\% \\
                               \cline{2-4}
                               &           \allCond & - & 0.812 & 48.568\% \\
\thline
\end{tabu}


%% file: 5_2.tex
\vspace{-0.5cm}
\subsection{Population Generalizability}  \label{populationGeneralizability}


The Population Generalizability experiment analyzed the algorithm's ability to estimate speech workload accurately on previously unseen individuals addressing question \R{2}. Estimating speech workload accurately for unseen individuals is necessary to support deployments in uncertain and dynamic environments. The hypotheses, denoted with a $2$ superscript, are identical to the prior experiment (Section \ref{emulatedRealWorld}).

\subsubsection{Methodology}

A leave-one-participant-out cross-validation method \cite{zhang2014crossvalidation} analyzed the algorithm's population generalizability using the Supervisory Evaluation results (Section \ref{supervisoryEval}). The algorithm was trained using 29 participant's data and tested using the remaining participant. This process was repeated 30 times, such that each participant was left out of the training once. The accuracy metrics were calculated using the aggregate testing data.

\subsubsection{Results}

The RMSE results were also provided in Table \ref{populationGeneralizabilityCorrelationRMSE} by workload condition, where the RMSE values range from 0.36 to 0.65 in the filtered dataset, and from 0.98 to 2.43 in the unfiltered dataset. The percent errors in the unfiltered dataset are all above 200\%, indicating a large degree of disagreement between the algorithm's estimates and model predictions. Conversely, the percent errors in the filtered dataset are around 12\% and 23\% in the OL and NL conditions, respectively, indicating a low disagreement. The filtered, UL condition's 113\% error reveals a moderate amount of disagreement.


The Pearson's Correlation coefficients are presented in Table \ref{populationGeneralizabilityCorrelationRMSE} by workload condition.The correlation across all workload conditions (All condition) in the unfiltered data set is positive and highly significant, but not all within condition (UL, NL, and OL) correlations are significant. The OL condition in the unfiltered dataset has a negative correlation significant correlation ($p<0.05$), while all other correlations are positive.

\begin{table}[h]
\caption{\correlationRMSECaption}
\input{Results/populationGeneralizability/correlationRMSE}
\label{populationGeneralizabilityCorrelationRMSE}
\end{table}

\subsubsection{Discussion}

Determining the algorithm's ability to generalize to unseen individuals is critical for deploying the estimated speech workload algorithm. 
The RMSE demonstrates the overall accuracy of the algorithm's estimate compared to the model prediction, which \h{RMSE}{2} predicted will be less than 5\% overall. However, this hypothesis was not supported. The leave-one-participant-out cross-validation paradigm may have caused the high error. The speech workload estimation algorithm was unable to train on the nuances of an individual's speech pattern, which presents a challenge for achieving a high population generalizability. However, the algorithm achieved moderate error for the filtered NL and OL conditions, indicating some population generalizability for those conditions. The algorithm's generalizability may be improved by using a long-short term memory network to capture temporal dynamics. 


A correlation assessed the strength of the algorithm's estimated response to changes in speech workload. \h{PC}{2}, which stated that the Pearson's correlation coefficient between the IMPRINT Pro model predictions and speech workload estimates will be positive and significant, was partially supported. The hypothesis was supported for all workload conditions using the filtered dataset, which indicates that the algorithm may accurately generalize to unseen individuals. The lack of complete support may be caused by algorithmic errors, or by IMPRINT Pro not modeling speech workload accurately.

This experiment did not consider differences between the training task environment and the real-world deployment. Thus, a cross-interaction paradigm analysis was conducted.


%% file: Results/populationGeneralizability/correlationRMSE.tex

\centering
\begin{tabu}{cc|llr}
\thline
\textbf{Dataset}               & \textbf{Condition} &   \textbf{Correlation} &   \textbf{RMSE} &   \textbf{Percent Error} \\
\hline
\multirow[c]{4}{*}{Unfiltered} &                 UL & 0.091** & 0.979 & 398.384\% \\
                               &                 NL & 0.008 & 1.748 & 372.709\% \\
                               &                 OL & -0.01* & 2.427 & 255.429\% \\
                               \cline{2-4}
                               &           \allCond & 0.077** & 1.776 & 330.119\% \\
\hline
\multirow[c]{4}{*}{Filtered}   &                 UL & - & 0.653 & 112.957\% \\
                               &                 NL & - & 0.268 & 22.805\% \\
                               &                 OL & - & 0.360 & 12.162\% \\
                               \cline{2-4}
                               &           \allCond & - & 0.460 & 25.346\% \\
\thline
\end{tabu}


%% file: 5_3.tex
\subsection{Human-Robot Teaming Paradigm Generalizability} \label{teamParadigmGenExp}

This experiment analyzed the speech workload estimation algorithm's generalizability across human-robot teaming paradigms and answers research question \R{3}. 
The hypotheses, denoted with a superscript $3$, are identical to  the Emulated Real-World Conditions experiment (Section \ref{emulatedRealWorld}).

\subsubsection{Methodology}

The analysis was performed via two-fold cross-validation. The first fold trained the algorithm on the data from the Peer-based evaluation (Section \ref{peerBasedEval}), and tested on the Supervisory Evaluation's data (Section \ref{supervisoryEval}). The second fold reversed the data sets for training and testing.

\subsubsection{Results}






The RMSE analyzes the estimate's accuracy, shown in Table \ref{humanRobotPeerCorrelationRMSE} for training on the supervisory evaluation and testing on the peer-based evaluation. Overall, the RMSE was somewhat higher across all conditions than the previous analyses. This result is attributed to the more unstructured setting of the peer-based evaluation, as dynamic environments elicit larger errors in the a priori IMPRINT Pro workload models. The Pearson's Correlation Coefficient for the unfiltered dataset are all at, or below 0.065 indicating difficulty in tracking workload across human-robot teaming paradigms. 

\begin{table}[!h]
\caption{\correlationRMSECaption}
\label{humanRobotPeerCorrelationRMSE}
\input{Results/humanRobotPeer/correlationRMSE}
\end{table}



The second result set focuses on the algorithm's generalizability from a peer-team to a supervisory-team. The RMSE values for the unfiltered and filtered datasets are similar to previous analyses indicating that the algorithm has less difficulty going from a peer-based paradigm to a supervisory-based one. However, the unfiltered, OL correlation coefficient is negative, indicating an inverse relationship between the IMPRINT Pro model's values and algorithm estimates. All other coefficients are positive and significant. The coefficients in the unfiltered dataset are all less than 0.12, indicating a mismatch between the model and estimates. 

\begin{table}[h]
\caption{\correlationRMSECaption}
\label{humanRobotSupervisoryCorrelationRMSE}
\input{Results/humanRobotSupervisory/correlationRMSE}
\vspace{-0.5cm}
\end{table}





\vspace{0.5cm}
\subsubsection{Discussion}

The goal of this experiment was to assess the algorithm's estimation accuracy using novel test data from a different interaction paradigm than the training data. The estimated accuracy with respect to the corresponding model prediction was expected, \h{RMSE}{3}, to result in a RMSE of less than 5\% overall, but this hypothesis was not supported. The large error in the filtered dataset from the Supervisory Evaluation was most likely due to a lack of UL speech workload data in the Peer-based Evaluation, evidenced by the model predictions in the Low workload condition from the Peer-based data and the UL condition in the Supervisory data.



\h{PC}{3} evaluated the algorithm's ability to track speech workload accurately when deployed in a different human-robot teaming paradigm than the one on which it was trained. This hypothesis was partially supported, which is attributed to the errors in the IMPRINT Pro models, as the filtered dataset correlations were strong. Errors occurred when there were high speech workload estimates as the participant spoke, but the IMPRINT Pro model did not expect speech. The opposite arose when IMPRINT Pro expected speech, but the participant was silent. A large number of instances where the IMPRINT Pro model prediction and speech workload estimate diverged may have confounded the correlation analysis.



%% file: Results/humanRobotPeer/correlationRMSE.tex

\centering
\begin{tabu}{cc|llr}
\thline
\textbf{Dataset}               & \textbf{Condition} &   \textbf{Correlation} &   \textbf{RMSE} &   \textbf{Percent Error} \\
\hline
\multirow[c]{3}{*}{Unfiltered} &                 Low & 0.014* & 2.330 & 216.129\% \\
                               &                 High & 0.065** & 1.992 & 209.718\% \\
                               \cline{2-5}
                               &           \bothCond & 0.040** & 2.155 & 213.553\% \\
\hline
\multirow[c]{3}{*}{Filtered}   &                 Low & - & 0.701 & 41.234\% \\
                               &                 High & - & 1.022 & 80.246\% \\
                               \cline{2-5}
                               &           \bothCond & - & 0.892 & 60.885\% \\
\thline
\end{tabu}


%% file: Results/humanRobotSupervisory/correlationRMSE.tex

\centering
\begin{tabu}{cc|llr}
\thline
\textbf{Dataset}               & \textbf{Condition} &   \textbf{Correlation} &   \textbf{RMSE} &   \textbf{Percent Error} \\
\hline
\multirow[c]{4}{*}{Unfiltered} &                 UL & 0.117** & 0.959 & 350.762\% \\
                               &                 NL & 0.005 & 1.641 & 353.042\% \\
                               &                 OL & -0.015* & 2.238 & 234.178\% \\
                               \cline{2-4}
                               &           \allCond & 0.073** & 1.658 & 301.752\% \\
\hline
\multirow[c]{4}{*}{Filtered}   &                 UL & - & 0.697 & 112.918\% \\
                               &                 NL & - & 0.509 & 44.853\% \\
                               &                 OL & - & 0.828 & 27.807\% \\
                               \cline{2-4}
                               &           \allCond & - & 0.726 & 40.484\% \\
\thline
\end{tabu}


%% file: 5_6.tex
\subsection{Physiological Data and Filler Utterances}  \label{physioFillerExp} 

The goal of the Physiological Data and Filler Utterances experiment was to assess the performance change resulting from adding features to supplement the base audio features described in Section \ref{algorithmDesign}. The population-generalizability analysis was performed (leave-one-participant-out cross validation) with respiration rate, filler utterances, and the features combined, which permits a direct performance comparison.
If either supplementary feature provides an increase in performance, the algorithm needs to include that feature to improve the speech workload estimate accuracy.
This experiment answers the research question (\R{4}): \emph{\featureSetQuestion{}}

The Pearson's correlation coefficient hypothesis, \h{PC}{4}, predicted an increase when the filler utterance and respiration rate features are included.
The second hypothesis, \h{RMSE}{4}, stated that the RMSE of the algorithm's estimates, when compared to the IMPRINT Pro speech workload predictions, will decrease when respiration rate and filler utterances are added.

\subsubsection{Methodology}

This experiment was conducted via a meta-analysis of the experiments presented in Section 5.2. The feature sets from all prior experiments were augmented with respiration rate, filler utterances, and both respiration rate and filler utterances separately.

\subsubsection{Results}

The accuracy metrics for determining algorithmic performance were calculated for each set of additional features: None (only the base audio features), Filler Utterances, Respiration Rate, and Both (respiration rate and filler utterances). The RMSE results are presented in Table \ref{featureSetRMSE}. The error deltas resulting from adding respiration rate, filler utterances, and both additional features for the unfiltered dataset were .011, .003, and .010, respectively, in the All condition. Including respiration rate, filler utterances, and both in the filtered dataset lead to decreases in error of .004, .006, and .005, respectively. The errors were lower when respiration was included in all conditions for both datasets. Including filler utterances did not affect the error in the overload condition for the unfiltered dataset, but did lower the error in all other cases. Including both additional features lead to the lowest errors in the overload condition for both datasets, but in all other conditions, the error was between the errors produced using each feature individually.

\begin{table}
\caption{RMSE of the Algorithm Estimates Compared to the IMPRINT Pro Model Speech Workload Predictions by Data Set, Workload Condition, and Additional Feature(s) Included.}
\input{Results/filledPausePhysio/realTimeWindowSize/RMSE}
\label{featureSetRMSE}
\vspace{-0.5cm}
\end{table}


The algorithm's accuracy tracking changes in speech workload was assessed using a correlation analysis, presented in Table \ref{featureSetCorrelation}. The correlation coefficients are all positive and significant, and the magnitudes increase with the additional features. The overall correlation coefficient in the unfiltered dataset increases when respiration rate and filler utterances are included. Augmenting the feature set with both respiration rate and filler utterances lead to an increased overall correlation coefficient in the unfiltered dataset, which is the same increase achieved with only respiration rate. The inclusion of respiration rate generally resulted in the highest correlation value in the unfiltered dataset. The inclusion of both additional features generally lead to a coefficient that was between the correlation coefficients produced by each feature independently.

\begin{table}
\caption{Correlation Between the Algorithm Estimates and the IMPRINT Pro Model Workload Predictions by Data Set, Workload Condition, and Additional Feature(s) Included.}
\label{featureSetCorrelation}
\input{Results/filledPausePhysio/realTimeWindowSize/correlation}
\end{table}



\subsubsection{Discussion}

The goal of the Physiological Data and Filler Utterances experiment was to determine the effect of adding respiration rate and filler utterances as features, which was quantified via Pearson's correlation coefficient and RMSE. The additional features resulted in marginally higher correlation coefficients over the base set of audio features, which partially supports \h{PC}{3}. Respiration rate and filler utterances are sensitive to changes in speech workload, which allowed the algorithm to make slightly more accurate estimates that tracked the model predictions. However, the additional features may not contain useful information that is not already encapsulated in the base feature set. Increased respiration rate leads to higher speech intensity and pitch, which are already in the base features.  Intensity and pitch may be influenced by stress, independent of changes in respiration rate. Filler utterances may not be a strong predictor of speech workload, and the relationship between speech workload and filler utterances may be inconsistent across individuals (i.e., some individuals may use more filler utterances when underloaded and omit when overloaded, or vice-versa). The additional features may improve accuracy by providing redundancy, resulting in an algorithm that is less susceptible to performance decrements when a metric is unavailable or noisy.

The instantaneous error when including respiration rate and filler utterances almost always resulted in minimal decreases in RMSE compared to the base set of audio features, partially supporting \h{RMSE}{3}. Respiration rate may have led to additional improvements due to the feature's accuracy compared to filler utterances. The respiration rate data was recorded via the BIOPAC\textsuperscript{\textregistered} Bioharness\textsuperscript{TM}, which can provide more accurate measurements. Few algorithms exist in the literature for detecting filler utterances, and the developed filler utterance detection algorithm has several limitations. The filler utterance detection algorithm assumes filler utterances are longer than 250ms, involve stable formants, and are comprised of back vowel sounds (e.g., ``uh'', ``oh'', ``oo''), which can lead to false negatives when assumptions are violated, or false positives when the speaker strings together several similar sounding syllables without performing glottal stops. Furthermore, filler utterances are not always present in speech. Redundant metrics are valuable even if they do not dramatically increase performance, because they can support the algorithm's estimation accuracy when one of the metrics is unavailable or noisy.

%% file: Results/filledPausePhysio/realTimeWindowSize/RMSE.tex

\centering
\begin{tabu}{X[l, 0.7]X[c, 0.8]|X[0.7,c]X[c]X[c]X[0.7,c]}

\thline
                                &      & \multicolumn{4}{c}{\textbf{Additional Feature(s) Included}} \\
                                       \cline{3-6}

\textbf{Dataset} &  \textbf{Condition} & \textbf{None} &  \textbf{Respiration Rate} &  \textbf{Filler Utterances} &  \textbf{Both}  \\
\hline

\multirow[c]{4}{*}{Unfiltered}  &         UL &  0.487 &  0.477 &  0.477 &  0.477 \\
                                &         NL &  1.113 &  1.089 &  1.111 &  1.091 \\
                                &         OL &  1.656 &  1.655 &  1.656 &  1.654 \\
                                \cline{2-6}
                                &   \allCond &  1.199 &  1.188 &  1.196 &  1.189 \\

\hline
\multirow[c]{4}{*}{Filtered}    &         UL &  0.270 &  0.261 &  0.263 &  0.260 \\
                                &         NL &  0.600 &  0.595 &  0.590 &  0.595 \\
                                &         OL &  0.735 &  0.732 &  0.733 &  0.731 \\
                                \cline{2-6}
                                &   \allCond &  0.581 &  0.577 &  0.575 &  0.576 \\
\thline
\end{tabu}

%% file: Results/filledPausePhysio/realTimeWindowSize/correlation.tex

\centering
\begin{tabu}{X[l, 0.7]X[c, 0.8]|X[0.7,c]X[c]X[c]X[0.7,c]}

\thline
                                &      & \multicolumn{4}{c}{\textbf{Additional Feature(s) Included}} \\
                                       \cline{3-6}

\textbf{Dataset} &  \textbf{Condition} & \textbf{None} &  \textbf{Respiration Rate} &  \textbf{Filler Utterances} &  \textbf{Both}  \\
\hline

\multirow[c]{4}{*}{Unfiltered}  &         UL &  0.180** &  0.182** &  0.183** &  0.181** \\
                                &         NL &  0.258** &  0.273** &  0.260** &  0.271** \\
                                &         OL &  0.213** &  0.216** &  0.215** &  0.216** \\
                                \cline{2-6}
                                &   \allCond &  0.330** &  0.337** &  0.332** &  0.337** \\
\thline
\end{tabu}






%% file: 5_Discussion.tex
\section{Overall Discussion}

The primary research objective underpinning all of the experiments was demonstrating the algorithm's ability to estimate speech workload accurately. The results demonstrate that the algorithm is capable of generating accurate estimates for unseen data. The accuracy metrics from the Emulated Real-World Conditions experiment (Section \ref{emulatedRealWorld}) were the highest, as training and testing used the same individuals, working in the same stationary supervisory-based human-robot teaming paradigm. The algorithm achieved high accuracy metrics when the participant, interaction paradigm, or task environment changed, indicating robustness to many variables that may be unknown during algorithm training.

The relatively high accuracy in the Population Generalizability experiment (Section \ref{populationGeneralizability}) was most likely due to the testing and training data being from the same evaluation. The algorithm was trained on all variables, except the validation set's left-out participant, meaning any over-fitting to the interaction paradigm aided accuracy and was undetected. The Human-Robot Teaming Paradigm experiment (Section \ref{teamParadigmGenExp}) included distinct individuals for distinct teaming relationships in the validation data, which may explain the lower accuracy. 

The IMPRINT Pro models' deficiencies were present throughout. These models were created prior to each evaluation, relied upon the specified static task start and end timing, and the resulting predictions did not adapt to the current experiment context. As a result, the models were unaware of the situations that arose during the human subjects evaluations that lead to a mismatch between the model's predictions and the human's actual speech activities and workload. The filtered datasets always displayed vastly superior accuracy metrics to the unfiltered datasets. It will be necessary to improve workload modeling capabilities in order to permit high accuracy and confidence in speech workload estimation. The metrics for the filtered dataset overestimate the algorithm's accuracy if there are errors in the voice activity algorithm.

There is an argument that the supervisory evaluation relied on pre-defined responses, which can decrease the generalizability of the algorithm. However, spurious speech did occur in the evaluation and the peer-based evaluation represents a less structured domain. Thus, the algorithm's generalizability is minimally impacted by the pre-defined verbal responses. However, future work will investigate domains where only spurious speech is used. The algorithm's generalizability may be enhanced as well by scaling the IMPRINT Pro model's values by the subjective ratings. This is left to future work, as such an approach needs to be validated and scaling IMPRINT Pro models does not occur in the literature.

Lastly, the algorithm is constrained to the english language. Other languages may exhibit differently in the various features used in the algorithm. For example, inflections are more common in french to differentiate words. This inflection will likely impact the pitch feature, which then impacts the workload estimate. It is also important to investigate continual learning approaches to fine-tune the algorithm to how an individual exhibits various speech workload levels.


%% file: 6_Conclusion.tex
\section{Conclusion}

Multi-dimensional workload estimates are critical for intelligent human-robot teaming systems. A system capable of adapting interactions based on the human's workload state can increase performance and reduce errors by mitigating undesirable workload states (overload and underload). Such a system will require a means of estimating the human's overall workload and each workload component to understand how to alter interaction modalities optimally in order to normalize the human's workload. The presented speech workload estimation algorithm was validated across multiple human subjects experiments that contained spurious speech. The algorithm greatly extends current speech workload estimation approaches by being invariant across individuals, human-robot teaming paradigms, and task environments, while being feasible for real-time deployment. Thus, the algorithm provides a robust solution for the speech workload estimation component necessary to achieve adaptive human-robot teams.

%% file: appendix.tex
\subsection{Peer-Based Evaluation Figures}
\label{appendix:peer}

\begin{figure}[!h]
    \centering
    \includegraphics[width=0.5\linewidth]{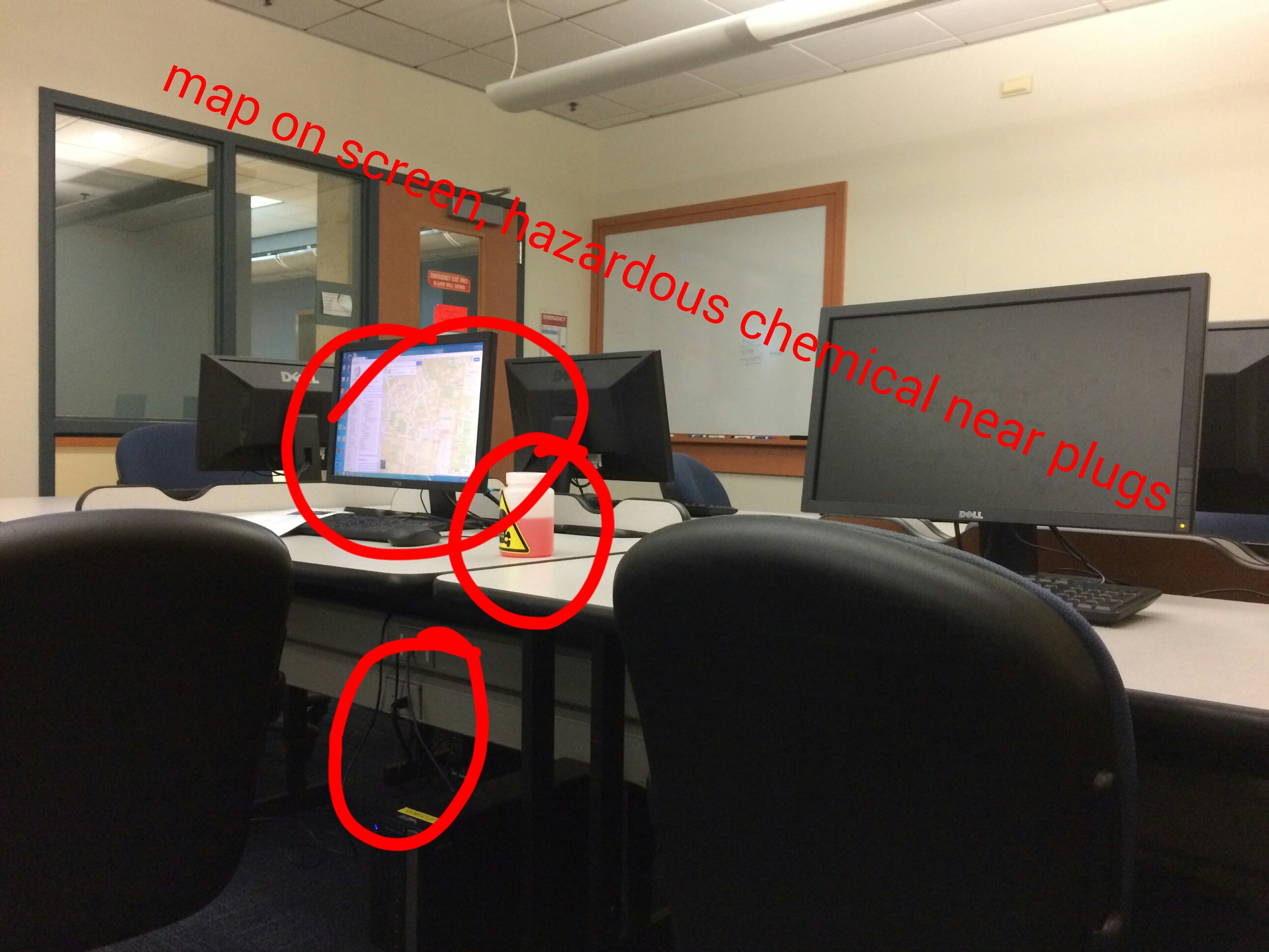}
    \caption{The peer-based evaluation's Task 1: suspicious items.}
    \label{fig:t1}
\end{figure}

\begin{figure}[!h]
    \centering
    \includegraphics[width=0.5\linewidth]{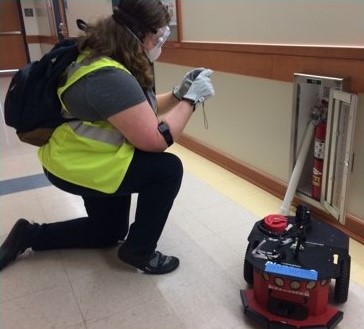}
    \caption{he peer-based evaluation's Task 2: materials search.}
    \label{fig:t2}
\end{figure}

\begin{figure}[!h]
    \centering
    \includegraphics[width=0.5\linewidth]{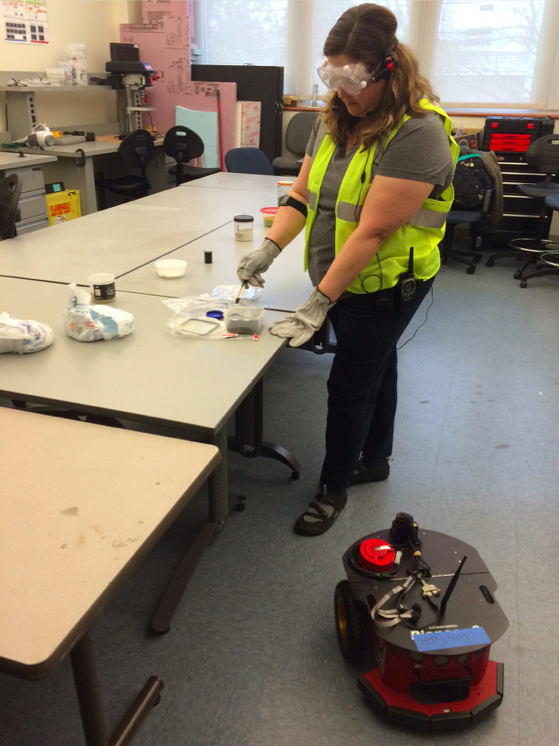}
    \caption{The peer-based evaluation's Tasks 3 and 4: sampling.}
    \label{fig:t34}
\end{figure}

\subsection{Descriptive Statistics}
\label{appendix:stats}

The following tables contain the descriptive statistics for each analysis in the main document. The simulated real-world conditions results are provided in Table \ref{realWorldDescriptive}. The estimated and model predicted means are within one standard deviation of each other across all conditions in both datasets. The algorithm estimated means and Std. Devs. increase as the workload level increases for the unfiltered and filtered datasets. A similar trend occurred with the filtered dataset; however, the algorithm's estimates were lower in the NL condition than the UL condition. The algorithm estimates during the UL condition underestimated speech workload for the unfiltered dataset, but overestimated speech workload for the filtered dataset.
The algorithm's maximum speech workload estimates ranged from 4.086 to 4.623 and are consistently greater than the maximum modeled workload predictions.

\begin{table}[!htb]
\caption{Simulated Real-World Conditions: Descriptive statistics by algorithm estimates and model predictions, data set and workload condition. \textbf{Note:} unless otherwise stated, all minimums for all results are 0.}
\input{Results/realWorld/descriptive}
\label{realWorldDescriptive}
\end{table}

The descriptive statistics for the algorithm's estimates and the IMPRINT Pro model predictions are provided in Table \ref{populationGeneralizabilityDescriptive} for the population generalizability analysis (Section \ref{populationGeneralizability}). The algorithm's estimates and model's predicted means are within one standard deviation of each other across all conditions for the both datasets. The model and algorithm's means increase as the workload level increases in both datasets, and the means are all higher in the filtered dataset. 

\begin{table}[!h]
\caption{Population Generalizability: Descriptive Statistics for the Algorithm Estimates
and the Speech Workload Model Predictions by
Data Set and Workload Condition.}
\input{Results/populationGeneralizability/descriptive}
\label{populationGeneralizabilityDescriptive}
\end{table}

The descriptive statistics for training on the supervisory evaluation and testing on the the peer evaluation (Section \ref{teamParadigmGenExp}) are presented in Table \ref{humanRobotPeerDescriptive}. The algorithm's means are consistently higher than the IMPRINT Pro means for the filtered data set, but are always lower than the model's for the unfiltered dataset. The means for the algorithm's estimates and IMPRINT Pro predictions are mostly within 0.15 of one another, except in the Low and Both conditions in the filtered dataset. 

\begin{table}[!h]
\caption{Peer-Based Evaluation: Descriptive Statistics for the Algorithm Estimates
and the Speech Workload Model Predictions by
Data Set and Workload Condition.}
\input{Results/humanRobotPeer/descriptive}
\label{humanRobotPeerDescriptive}
\end{table}

The resulting descriptive statistics for training on the peer-based evaluation ant testing on the supervisory-based evaluation (Section \ref{teamParadigmGenExp}) are presented in Table \ref{environmentGeneralizabilityDescriptive}. The algorithm means are consistently lower than the model prediction means for the unfiltered dataset. The filtered dataset's UL condition's algorithm mean is greater than the model's, while the remaining conditions all resulted in lower algorithm means than model's. Further, the algorithm means are similar to the IMPRINT Pro means, with differences around 0.1, while the difference between the means in the filtered dataset range from 0.2--0.7. Both sets of results increase as workload increases. 

\begin{table}
\caption{Supervisory-Based Evaluation: Descriptive Statistics for the Algorithm Estimates
and the Speech Workload Model Predictions by
Data Set and Workload Condition.}
\input{Results/humanRobotSupervisory/descriptive}
\label{environmentGeneralizabilityDescriptive}
\end{table}

\subsection{Algorithm Run-Time}
\label{appendix:run}

The window-size chosen can impact algorithm run-time significantly, due to the inclusion of spectral/frequency-based features. Table \ref{windowSizeRunTime} provides the results of running the algorithm ten times on the same computer using CPU resources (2013 HP desktop computer running Ubuntu with a dual core Intel Core i3 (3.4 GHz) processor and 6GB of RAM). Overall, the algorithm ran in linear (O(n)) time, meaning that the total run-time increased in a constant fashion as the window size increased. The total runtime was less than one second when the window size was 15 seconds or less, which means that speech workload can be estimated every second. Pitch feature extraction was the largest component of the total run-time, while the intensity feature was the smallest contributor.

\begin{table}[!h]
\caption{The Mean (Std. Dev.) Algorithm Run-Time for Each Feature and Window Size. Run-times were measured in Seconds.}
\label{windowSizeRunTime}
\input{Results/realTimeWindowSize/runTime}
\end{table}

\begin{table*}[!t]
\centering
\small
\renewcommand{\arraystretch}{1.2}
\begin{tabular}{lccccccccccc}
\hline
\textbf{Validation Set} & \textbf{1} & \textbf{2} & \textbf{3} & \textbf{4} & \textbf{5} & \textbf{6} & \textbf{7} & \textbf{8} & \textbf{9} & \textbf{10} & \textbf{All} \\
\hline
Seconds Analyzed & 2610 & 2610 & 2610 & 2610 & 2610 & 2610 & 2610 & 2610 & 2610 & 2610 & 26100 \\
PRAAT Detections & 616 & 479 & 542 & 612 & 522 & 458 & 583 & 540 & 551 & 567 & 5470 \\
Algorithm Detections & 911 & 736 & 817 & 842 & 774 & 646 & 863 & 761 & 810 & 832 & 7992 \\
Correct Detections & 588 & 453 & 520 & 578 & 499 & 415 & 546 & 505 & 528 & 540 & 5172 \\
False Alarms & 323 & 283 & 297 & 264 & 275 & 231 & 317 & 256 & 282 & 292 & 2820 \\
Precision & 0.65 & 0.62 & 0.64 & 0.69 & 0.64 & 0.64 & 0.63 & 0.66 & 0.65 & 0.65 & 0.65 \\
Recall & 0.95 & 0.95 & 0.96 & 0.94 & 0.96 & 0.91 & 0.94 & 0.94 & 0.96 & 0.95 & 0.95 \\
F1 Score & 0.77 & 0.75 & 0.77 & 0.80 & 0.77 & 0.75 & 0.75 & 0.78 & 0.78 & 0.77 & 0.77 \\
\hline
\end{tabular}
\caption{Syllable detection performance across 10 validation sets and the aggregated total.}
\label{tab:syllable-validation}
\end{table*}

\begin{table*}[!t]
\centering
\small
\renewcommand{\arraystretch}{1.2}
\begin{tabular}{lccccccccccc}
\hline
\textbf{Validation Set} & \textbf{1} & \textbf{2} & \textbf{3} & \textbf{4} & \textbf{5} & \textbf{6} & \textbf{7} & \textbf{8} & \textbf{9} & \textbf{10} & \textbf{All} \\
\hline
Seconds Analyzed & 2610 & 2610 & 2610 & 2610 & 2610 & 2610 & 2610 & 2610 & 2610 & 2610 & 26100 \\
rVAD Voice Activity (s) & 654 & 609 & 585 & 601 & 611 & 605 & 603 & 602 & 629 & 588 & 6087 \\
Algorithm Voice Activity (s) & 348 & 315 & 318 & 303 & 286 & 341 & 303 & 311 & 313 & 297 & 3135 \\
Correct Detections & 271 & 249 & 265 & 262 & 237 & 272 & 233 & 250 & 268 & 224 & 2531 \\
False Alarms & 77 & 66 & 53 & 41 & 49 & 69 & 70 & 61 & 45 & 73 & 604 \\ \hline
Accuracy & 0.82 & 0.84 & 0.86 & 0.85 & 0.84 & 0.85 & 0.83 & 0.84 & 0.84 & 0.83 & 0.84 \\
Precision & 0.78 & 0.79 & 0.83 & 0.86 & 0.83 & 0.80 & 0.77 & 0.80 & 0.86 & 0.75 & 0.81 \\
Recall & 0.41 & 0.41 & 0.45 & 0.44 & 0.39 & 0.45 & 0.39 & 0.42 & 0.43 & 0.38 & 0.42 \\
F1 Score & 0.54 & 0.54 & 0.58 & 0.58 & 0.53 & 0.58 & 0.52 & 0.55 & 0.57 & 0.50 & 0.55 \\
\hline
\end{tabular}
\caption{Voice activity detection performance metrics across 10 validation sets and the aggregated total.}
\label{tab:vad-validation}
\end{table*}

\begin{table*}[!t]
\centering
\small
\renewcommand{\arraystretch}{1.2}
\begin{tabular}{lcccccccccccc}
\hline
\textbf{Participant} & \textbf{102} & \textbf{103} & \textbf{104} & \textbf{106} & \textbf{107} & \textbf{108} & \textbf{109} & \textbf{110} & \textbf{111} & \textbf{113} & \textbf{114} & \textbf{All} \\
\hline
Filler Utterances & 50 & 78 & 7 & 48 & 2 & 29 & 2 & 8 & 0 & 0 & 0 & 224 \\
Algorithm Detections & 36 & 9 & 6 & 30 & 19 & 15 & 1 & 1 & 6 & 32 & 1 & 156 \\
\hline
\end{tabular}
\caption{Detected and ground truth filler utterances across participants.}
\label{tab:filler-utterances}
\end{table*}

\subsection{Validation of Core Speech-Features}
\label{appendix:feats}

The voice activity detection algorithm was validated using random validation sets taken from the first day of the Supervisory Evaluation. Ten sets comprised of 30-second segments of audio were generated for each participant, in each workload condition of the evaluation. Each segment was created by choosing a random starting point in milliseconds and isolating the following 30 s of audio data.  The detections produced by the voice activity algorithm, before computing the summary statistics, were compared to the voice activity values produced by rVAD, and the number of true positives, false positives, true negatives, and false negatives were determined. The parameters of rVAD were tuned on an additional random validation set with human-transcribed ground-truth labels, such that rVAD attained the highest F1 score. The validation of the voice activity algorithm is presented in Table \ref{tab:syllable-validation}.

The syllable detection algorithm was validated using the same random validation sets produced for validating the voice activity algorithm. A PRAAT Script for Detecting Syllable Nuclei was employed for generating groundtruth syllable timestamps. The parameters of the PRAAT syllable algorithm were tuned to maximize F1 score on an additional validation set with human-transcribed ground-truth timestamps. The syllable detection algorithm’s timestamps, generated before the count of the total number of filled pauses was taken, were compared against the PRAAT syllable algorithm’s timestamps. Syllable detections that were with 0.1 s of a syllable detected by the PRAAT syllable algorithm were considered to be true positives, while all others were false positives. The precision, recall, and F1 score, shown in Table \ref{tab:vad-validation}, were calculated for the syllable detection algorithm’s timestamps

There was no published algorithm for detecting filler utterances that was publicly available, so the filler utterance detection algorithm was validated via a corpus containing filler \newpage utterances. The Crosslinguistic Corpus of Hesitation Phenomena (CCHP) includes audio from 11 participants engaged in various speech tasks, accompanied by human-transcriptions, which included filler utterances. Only filler utterances labeled “uh”, “um”, and “mm” in the corpus were considered, since these were the only filler utterances the algorithm was designed to detect. The comparison of the number of filler utterances spoken by each participant, and the corresponding number of filler utterances detected by the algorithm are presented in Table \ref{tab:filler-utterances}.

\newpage

%% file: Results/realWorld/descriptive.tex
\footnotesize
\centering
\begin{tabu}{ccc|llll}

\thline
  &  \textbf{Condition} & \textbf{Source} &  \textbf{Mean} &  \textbf{Std. Dev.} &  \textbf{Median} &  \textbf{Max.} \\
\hline
\multirow[c]{8}{*}{\rotatebox{90}{Unfiltered Dataset}} & \multirow[c]{2}{*}{UL}       &    Model &  0.511 &  0.872 &  0 &  2 \\
                               &                              & Algorithm &  0.253 &  0.974 &  0 &  4.153 \\
                               \cline{2-7}
                               & \multirow[c]{2}{*}{NL}       &    Model &  0.507 &  1.33 &  0 &  4 \\
                               &                              & Algorithm &  0.482 &  1.303 &  0 &  4.623 \\
                               \cline{2-7}
                               & \multirow[c]{2}{*}{OL}       &    Model &  1.004 &  1.734 &  0 &  4 \\
                               &                              & Algorithm &  1.131 &  1.802 &  0 &  4.316 \\
                               \cline{2-7}
                               & \multirow[c]{2}{*}{\allCond} &    Model &  0.666 &  1.332 &  0 &  4 \\
                               &                              & Algorithm &  0.587 &  1.416 &  0 &  4.623 \\
\hline
\multirow[c]{8}{*}{\rotatebox{90}{Filtered Dataset}}   & \multirow[c]{2}{*}{UL}       &    Model &  0.836 &  0.986 &  0 &  2 \\
                               &                              & Algorithm &  1.673 &  1.975 &  0 &  4.145 \\
                               \cline{2-7}
                               & \multirow[c]{2}{*}{NL}       &    Model &  1.179 &  1.824 &  0 &  4 \\
                               &                              & Algorithm &  1.179 &  1.824 &  0 &  4.086 \\
                               \cline{2-7}
                               & \multirow[c]{2}{*}{OL}       &    Model &  2.742 &  1.857 &  4 &  4 \\
                               &                              & Algorithm &  2.743 &  1.858 &  3.999 &  4.119 \\
                               \cline{2-7}
                               & \multirow[c]{2}{*}{\allCond} &    Model &  1.672 &  1.799 &  1 &  4 \\
                               &                              & Algorithm &  2.001 &  2.001 &  1.992 &  4.145 \\
\thline
\end{tabu}

%% file: Results/populationGeneralizability/descriptive.tex

\centering
\begin{tabu}{ccc|llll}
\thline
 &  \textbf{Condition} & \textbf{Source} &  \textbf{Mean} &  \textbf{Std. Dev.} &  \textbf{Median} &  \textbf{Max.} \\
\hline
\multirow[c]{8}{*}{\rotatebox{90}{Unfiltered Dataset}} & \multirow[c]{2}{*}{UL}       &    Model &  0.246 &  0.657 &  0 &  2 \\
                               &                              & Algorithm &  0.188 &  0.786 &  0 &  5.414 \\
                               \cline{2-7}
                               & \multirow[c]{2}{*}{NL}       &    Model &  0.469 &  1.287 &  0 &  4 \\
                               &                              & Algorithm &  0.435 &  1.193 &  0 &  5.625 \\
                               \cline{2-7}
                               & \multirow[c]{2}{*}{OL}       &    Model &  0.95 &  1.702 &  0 &  4 \\
                               &                              & Algorithm &  1.103 &  1.707 &  0 &  4.37 \\
                               \cline{2-7}
                               & \multirow[c]{2}{*}{\allCond} &    Model &  0.538 &  1.291 &  0 &  4 \\
                               &                              & Algorithm &  0.556 &  1.322 &  0 &  5.625 \\
\hline
\multirow[c]{8}{*}{\rotatebox{90}{Filtered Dataset}}   & \multirow[c]{2}{*}{UL}       &    Model &  0.578 &  0.906 &  0 &  2 \\
                               &                              & Algorithm &  0.889 &  1.427 &  0 &  4.085 \\
                               \cline{2-7}
                               & \multirow[c]{2}{*}{NL}       &    Model &  1.177 &  1.823 &  0 &  4 \\
                               &                              & Algorithm &  1.082 &  1.688 &  0 &  4.086 \\
                               \cline{2-7}
                               & \multirow[c]{2}{*}{OL}       &    Model &  2.959 &  1.755 &  4 &  4 \\
                               &                              & Algorithm &  2.757 &  1.658 &  3.75 &  4.278 \\
                               \cline{2-7}
                               & \multirow[c]{2}{*}{\allCond} &    Model &  1.816 &  1.897 &  1 &  4 \\
                               &                              & Algorithm &  1.801 &  1.829 &  0.949 &  4.278 \\
\thline
\end{tabu}

%% file: Results/humanRobotPeer/descriptive.tex

\centering
\begin{tabu}{ccc|llll}
\thline
 &  \textbf{Condition} & \textbf{Source} &  \textbf{Mean} &  \textbf{Std. Dev.} &  \textbf{Median} &  \textbf{Max.} \\
\hline
\multirow[c]{6}{*}{\rotatebox{90}{Unfiltered Dataset}} & \multirow[c]{2}{*}{Low}       &    Model &  1.078 &  1.642 &  0 &  4 \\
                               &                              & Algorithm &  1.066 &  1.676 &  0 &  4.882 \\
                               \cline{2-7}
                               & \multirow[c]{2}{*}{High}       &    Model &  0.95 &  1.343 &  0 &  4 \\
                               &                              & Algorithm &  0.853 &  1.557 &  0 &  4.792 \\
                               \cline{2-7}
                               & \multirow[c]{2}{*}{\bothCond} &    Model &  1.009 &  1.491 &  0 &  4 \\
                               &                              & Algorithm &  0.952 &  1.617 &  0 &  4.882 \\
\hline
\multirow[c]{6}{*}{\rotatebox{90}{Filtered Dataset}}   & \multirow[c]{2}{*}{Low}       &    Model &  1.701 &  1.809 &  2 &  4 \\
                               &                              & Algorithm &  1.861 &  1.847 &  3.07 &  4.096 \\
                               \cline{2-7}
                               & \multirow[c]{2}{*}{High}       &    Model &  1.273 &  1.431 &  0 &  4 \\
                               &                              & Algorithm &  1.831 &  1.849 &  0 &  4.792 \\
                               \cline{2-7}
                               & \multirow[c]{2}{*}{\bothCond} &    Model &  1.465 &  1.626 &  1 &  4 \\
                               &                              & Algorithm &  1.845 &  1.848 &  1.42 &  4.792 \\
\thline
\end{tabu}

%% file: Results/humanRobotSupervisory/descriptive.tex

\centering
\begin{tabu}{ccc|llll}
\thline
 & \textbf{Condition} & \textbf{Source} &  \textbf{Mean} &  \textbf{Std. Dev.} &  \textbf{Median} &  \textbf{Max.} \\
\hline
\multirow[c]{8}{*}{\rotatebox{90}{Unfiltered Dataset}} & \multirow[c]{2}{*}{UL}       &    Model &  0.273 &  0.687 &  0 &  2 \\
                               &                              & Algorithm &  0.183 &  0.748 &  0 &  6.345 \\
                               \cline{2-7}
                               & \multirow[c]{2}{*}{NL}       &    Model &  0.465 &  1.282 &  0 &  4 \\
                               &                              & Algorithm &  0.376 &  1.026 &  0 &  7.476 \\
                               \cline{2-7}
                               & \multirow[c]{2}{*}{OL}       &    Model &  0.956 &  1.706 &  0 &  4 \\
                               &                              & Algorithm &  0.92 &  1.423 &  0 &  6.408 \\
                               \cline{2-7}
                               & \multirow[c]{2}{*}{\allCond} &    Model &  0.549 &  1.296 &  0 &  4 \\
                               &                              & Algorithm &  0.478 &  1.13 &  0 &  7.476 \\
\hline
\multirow[c]{8}{*}{\rotatebox{90}{Filtered Dataset}}   & \multirow[c]{2}{*}{UL}       &    Model &  0.618 &  0.924 &  0 &  2 \\
                               &                              & Algorithm &  0.981 &  1.488 &  0 &  5.628 \\
                               \cline{2-7}
                               & \multirow[c]{2}{*}{NL}       &    Model &  1.135 &  1.803 &  0 &  4 \\
                               &                              & Algorithm &  0.888 &  1.428 &  0 &  6.002 \\
                               \cline{2-7}
                               & \multirow[c]{2}{*}{OL}       &    Model &  2.977 &  1.745 &  4 &  4 \\
                               &                              & Algorithm &  2.296 &  1.369 &  3.053 &  5.821 \\
                               \cline{2-7}
                               & \multirow[c]{2}{*}{\allCond} &    Model &  1.794 &  1.883 &  1 &  4 \\
                               &                              & Algorithm &  1.555 &  1.574 &  1.379 &  6.002 \\
\thline
\end{tabu}

%% file: Results/realTimeWindowSize/runTime.tex

\centering
\begin{tabu}{X[1.3]|XXXXXX}

\thline
                        & \multicolumn{6}{c}{\textbf{Window Size (Seconds)}} \\
                        \cline{2-7}

\textbf{Feature}        &  \textbf{1s} &  \textbf{5s} &  \textbf{10s} &  \textbf{15s} &  \textbf{30s} &  \textbf{60s} \\
\hline

Intensity      &  .001 (.00) &  .007 (.00) &  .013 (.00) &  .019 (.00) &  .038 (.00) &  .069 (.01) \\
\hline

Pitch          &  .051 (.02) &  .246 (.08) &  .490 (.15) &  .734 (.22) &  1.46 (.44) &  2.84 (.88) \\
\hline

Voice Activity &  .004 (.00) &  .024 (.00) &  .049 (.00) &  .074 (.00) &  .149 (.00) &  .286 (.02) \\
\hline

Speech-Rate    &  .004 (.00) &  .024 (.00) &  .047 (.00) &  .070 (.00) &  .139 (.00) &  .258 (.03) \\
\hline

All Features   &  .061 (.02) &  .301 (.08) &  .599 (.15) &  .897 (.22) &  1.78 (.44) &  3.45 (.88) \\
\hline

\thline
\end{tabu}